\documentclass[letterpaper, 10 pt, conference]{ieeeconf}  

\IEEEoverridecommandlockouts                              

\usepackage{amsmath} 
\usepackage{amssymb}  
\usepackage{xurl}
\usepackage{graphicx}
\usepackage{placeins}
\usepackage{siunitx}
\usepackage{algorithm}
\usepackage{algpseudocode}
\usepackage{subfigure}
\usepackage{booktabs}
\usepackage{multirow}
\usepackage{dblfloatfix}
\usepackage{hyperref}
\usepackage{todonotes}
\usepackage{acronym}
\usepackage{siunitx}
\usepackage{bm}
\usepackage[capitalise]{cleveref}
\usepackage[perpage]{footmisc}

\acrodef{RL}{Reinforcement Learning}
\acrodef{NN}{Neural Network}
\acrodef{DoF}{Degree of Freedom}
\acrodef{ID}{Inverse Dynamics}
\acrodef{IMU}{Inertial Measurement Unit}
\acrodef{PPO}{Proximal Policy Optimization}
\acrodef{MLP}{Multi-Layer Perceptron}

\title{
Dynamic Throwing with Robotic Material Handling Machines
}

\author{Lennart Werner$^{*}$, Fang Nan$^{*}$, Pol Eyschen, Filippo A. Spinelli, Hongyi Yang, and Marco Hutter
\thanks{$^{*}$The authors contributed equally to this work.}%
\thanks{All authors are with Robotic Systems Lab, ETH Zurich, 8092 Zurich, Switzerland. \texttt{\{lwerne, fannan, peyschen, fspinelli, hoyang, mahutter\}@ethz.ch}}.
\thanks{This project has received funding from the European Union’s Horizon Europe Framework Programme under grant agreement No 101070405.}
}

\begin{document}

\maketitle
\thispagestyle{empty}
\pagestyle{empty}

\begin{abstract}

Automation of hydraulic material handling machinery is currently limited to semi-static pick-and-place cycles.
Dynamic throwing motions which utilize the passive joints, can greatly improve time efficiency as well as increase the dumping workspace.
In this work, we use \ac{RL} to design dynamic controllers for material handlers with underactuated arms as commonly used in logistics.
The controllers are tested both in simulation and in real-world experiments on a 12-ton test platform.
The method is able to exploit the passive joints of the gripper to perform dynamic throwing motions.
With the proposed controllers, the machine is able to throw individual objects to targets outside the static reachability zone with good accuracy for its practical applications.
The work demonstrates the possibility of using \ac{RL} to perform highly dynamic tasks with heavy machinery, suggesting a potential for improving the efficiency and precision of autonomous material handling tasks.

\end{abstract}

\section*{Supplymentary video}
A video summary of this work, including footages of simulation and real-world experiments, can be found at \url{https://youtu.be/YjvJTmGk0iI}.

\section{Introduction}
\label{sec:intro}

\begin{figure}[h]
  \centering
  \subfigure[Waste sorting]{
    \includegraphics[width=0.5\linewidth]{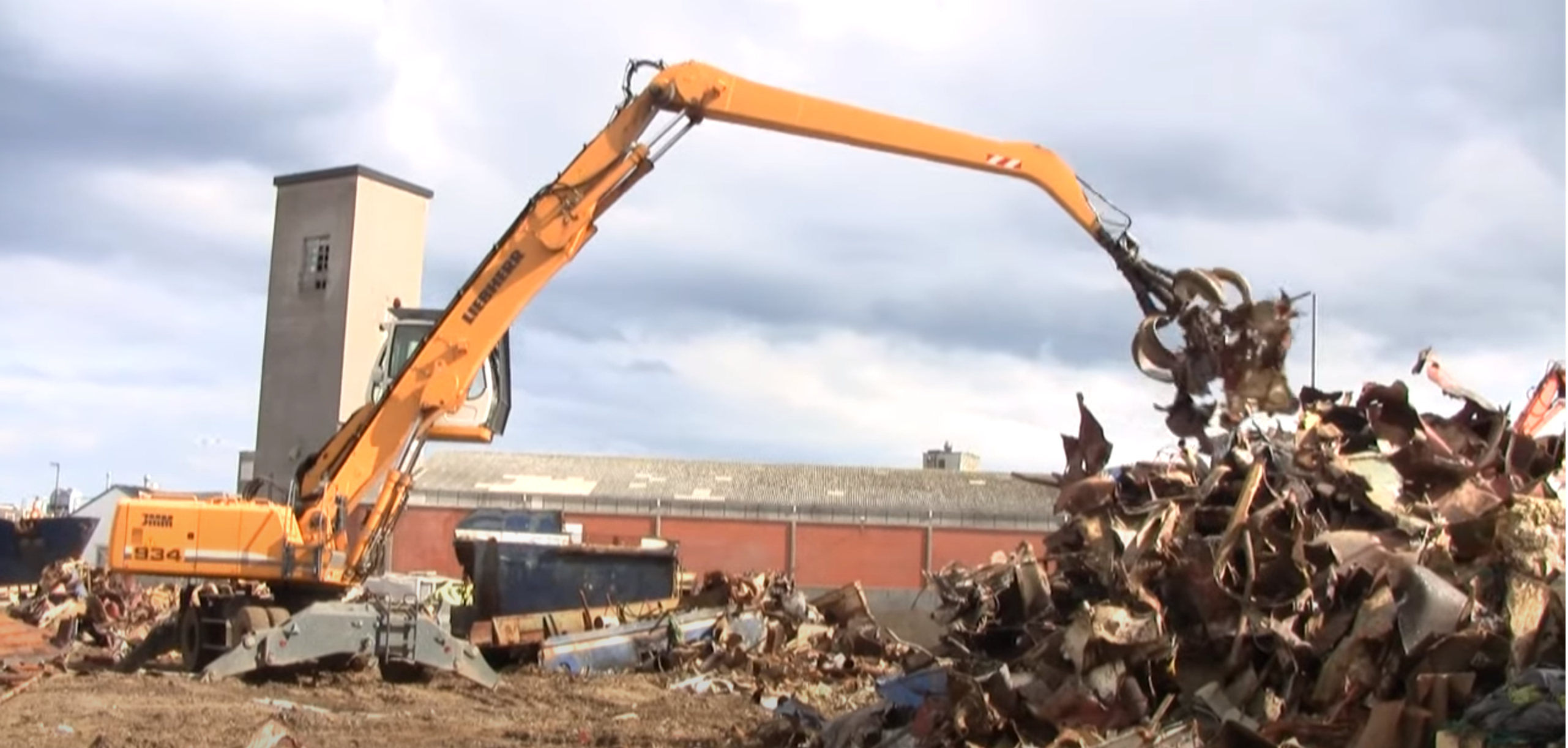}
  }
  \subfigure[Bulk material loading]{
    \includegraphics[width=0.43\linewidth]{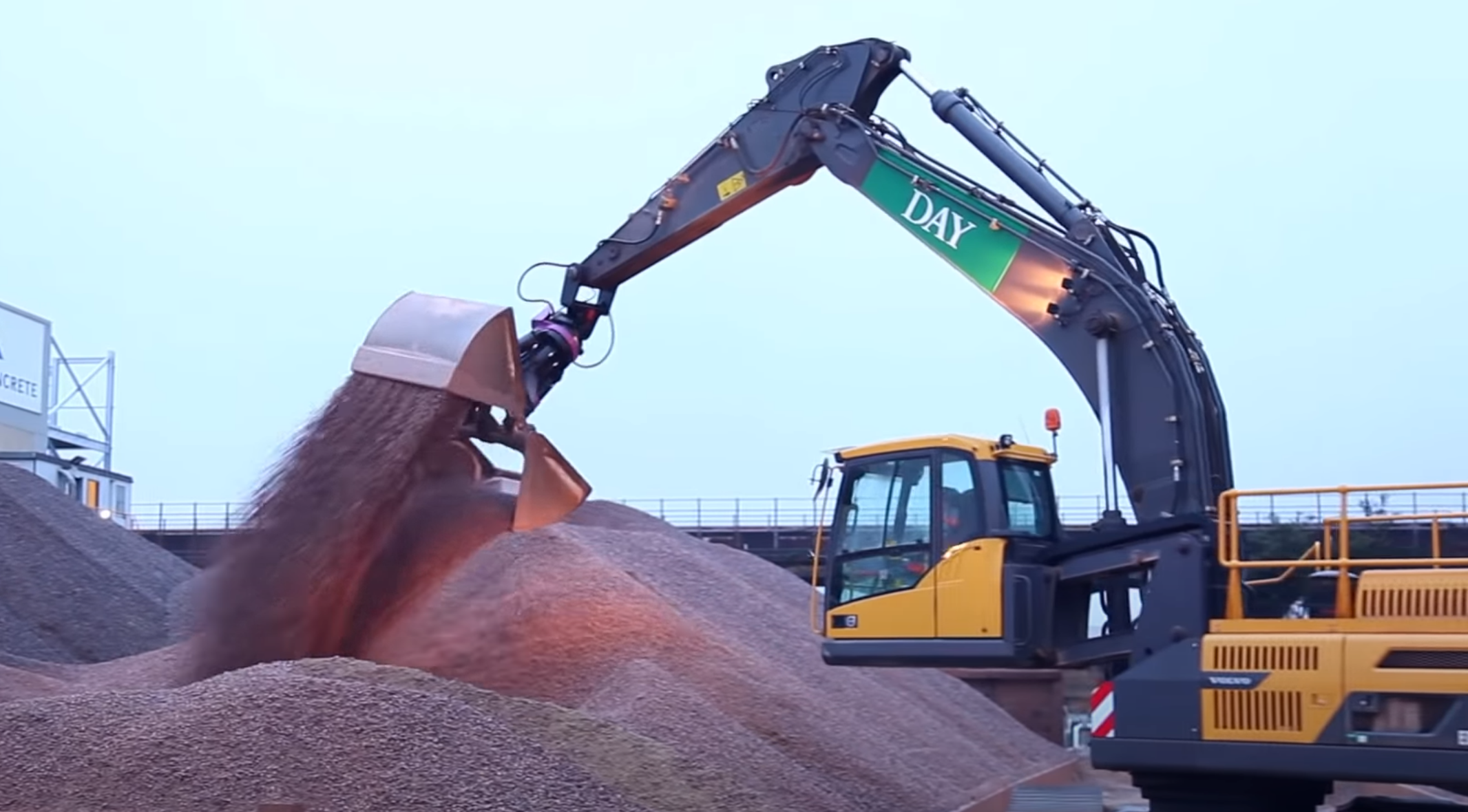}
  }
  \caption[Human operator throwing]{Human operators using material handling machines to perform throwing in different applications.\footnotemark }
  \label{fig:humanop_throwing}
\end{figure}

Automation of hydraulic machinery has recently gained increased attention in the research domain, driven by its relevance to construction, agriculture, forestry, mining or logistics~\cite{Johns23FrameworkRobotic,Jelavic22RoboticPrecision,Spinelli24ReinforcementLearning}. 
The advantages of automation in this context primarily consist of mitigating risks for human operators, enhancing accuracy, and providing a solution to address labor shortage. 
Yet, the performance of automated systems frequently lags behind that of skilled human operators across a range of tasks. 
Bulk material handling is efficiently performed by large machines equipped with hydraulic actuators and underactuated joints connecting the gripper to the arm. These machines are used for loading, unloading, or sorting tasks.
The underactuated design facilitates adaptation to various material shapes and sizes, enabling human operators to execute agile maneuvers that elevate operational efficiency. As illustrated in~\cref{fig:humanop_throwing}, human operators can exploit the dynamics of the freely-swinging gripper to throw payloads toward specific targets. This skill, which can be achieved by accurately coordinating minimal arm movement with the gripper oscillations, allows for the targeting of locations beyond the arm's conventional operational range.

Despite the benefits in efficiency and increased dumping workspace, performing such maneuvers precisely requires a lot of practice, resulting extremely challenging for inexperienced operators. 
A dynamic throwing motion is always associated with a high risk of self- and obstacle collisions.
Moreover, the intricate dynamics of the hydraulic actuation as well as the passive gripper pose significant challenges to the modeling process, thereby complicating task planning and control.
Even small control errors could be magnified by the passive joints and lead to very different end-effector trajectories.
Because of these complex challenges, automation of underactuated material handling machines has rarely been addressed in the past.
Previous work~\cite{Spinelli24ReinforcementLearning} presented an \ac{RL}-based approach to design an anti-swing controller for such machines. Drawing inspiration from it, our research  leverages the dynamics of the free-swinging gripper to learn dynamic handling skills, rather than suppressing the oscillation. 
Such an ability could reduce the cycle time, increase the operational space, reduce energy consumption, and therefore increase the efficiency of the overall process.

\begin{figure}[t]
  \centering
  \includegraphics[width = \linewidth]{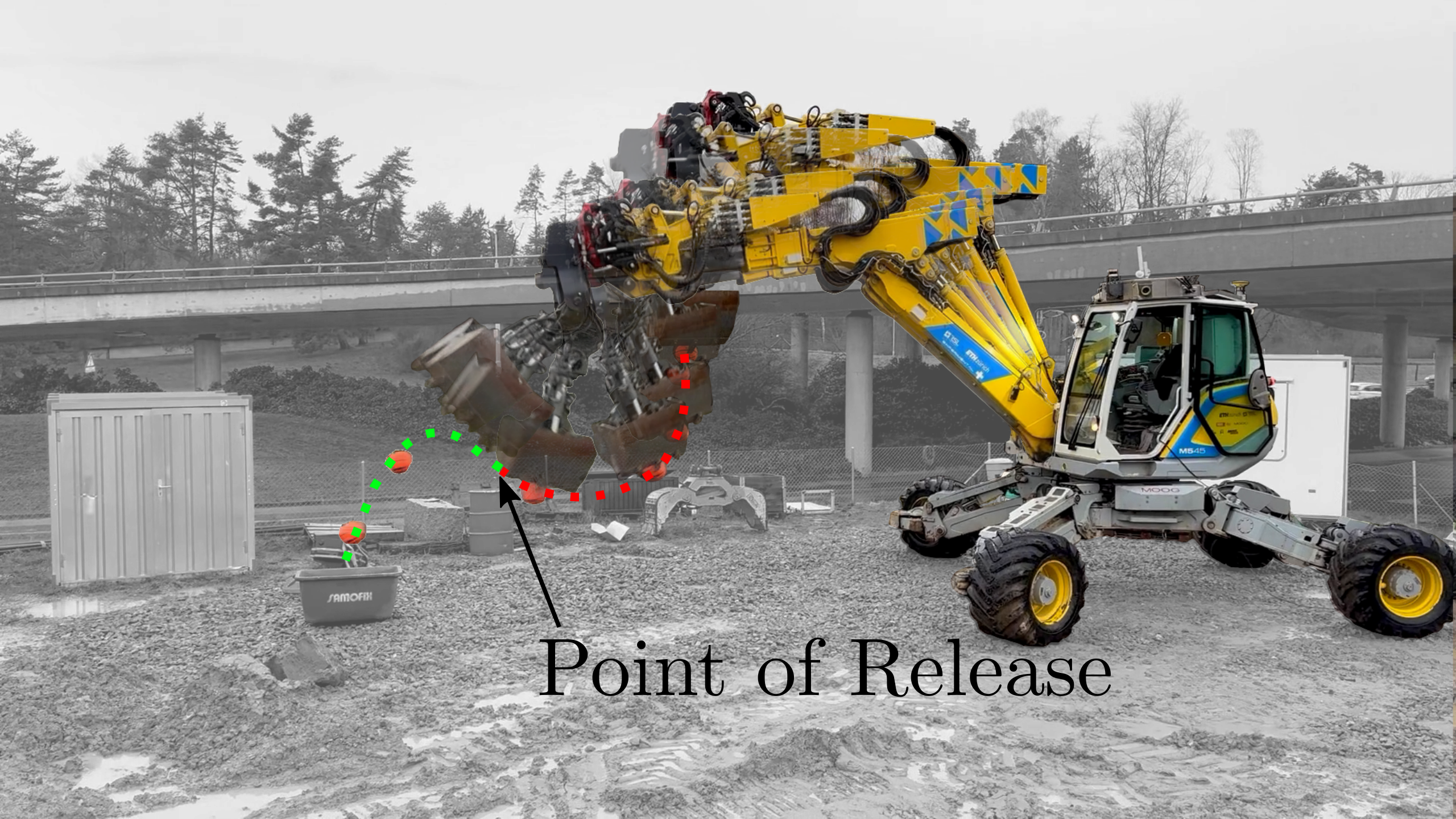}
  \caption{Two-dimensional throwing controller deployed on the M545 excavator with overlayed payload trajectory. }
  \label{fig:throwing_trajectory}
\end{figure}

\subsection{Contributions}
This work presents the first dynamic throwing controller for hydraulic material handling machinery.
The controller leverages the underactuated joints of material-handling machines to perform dynamic throwing motions, as illustrated in \cref{fig:throwing_trajectory}. 
With appropriate sim-to-real techniques, the policy, purely trained in simulation, was successfully deployed on a 12-ton material handling test platform.
An in-depth accuracy analysis benchmarks the synthesized controllers.
Real world experiments confirm the identified precision.
The experiments show that the controller is able to perform accurate, repeatable dumping of the payload in motion even outside of the regular machine workspace.

\footnotetext{Picture source: \url{https://www.youtube.com/watch?v=i46xzwaQB50}, \url{https://www.youtube.com/watch?v=KFxRoSshLGI}}
\section{Related Work}
\label{sec:related_work}
\subsection{Robot Throwing}
Manipulation has always been a key direction in robotic research, with many applications in factories, warehouses, and households~\cite{Billard19TrendsChallenges}.
Instead of static pick-and-place of objects, a number of research works focused on using robotic throwing to increase time and energy efficiency.
For example, ~\cite{Chen19RobotThrowing} presents a throwing trajectory planning algorithm for solid waste handling, taking into account robot kinematic constraints.
A later method plans throwing trajectories also taking into account the objects' flying dynamics~\cite{Liu22SolutionAdaptive}. Through fast online replanning, the system is robust under disturbance.
These works have not been implemented in a real-world experiment.

Besides classical model-based methods, learning-based approaches are also used for robotic throwing.
An early work is done in~\cite{Kober11LearningElementary}, which uses a hierarchical learning framework to train a planar robotic arm to throw balls at one of three fixed targets.
Inverse \ac{RL} has been used in~\cite{Loncarevic19LearningRobotic} to train a humanoid robot to throw balls at a target based on human feedback.
A learning-based method to predict a correction to the throwing velocity was presented in~\cite{Zeng20TossingBotLearning}, enabling the throwing of arbitrary objects.
Recently, imitation learning has also been used to learn throwing from human demonstrations~\cite{Chi24UniversalManipulation}.
Both the last two works showed a remarkable success rate in real-world experiments, but they relied on an existing arm controller to throw the object with the planned release velocity. Reaching a certain throwing velocity at a specific release position, however, is a non-trivial problem for underactuated robot manipulators, including material handling machines.

There have been some studies of throwing with an underactuated robotic arm, inspired mainly by ball-pitching motions of humans~\cite{Mettin10OptimalBall, Katsumata09ThrowingMotion}.
A similar problem has been studied in~\cite{Gai13MotionControl}, which focused on a robot arm with a flexible link.
The authors showed that flexibility helps the arm to perform fast motion within a short operation time.
All of these studies, however, use a model-based approach to design controllers that output torque commands.
In real-world applications, such models can hardly capture the full dynamics of the hardware.
For this reason, some works address modeling with learning methods~\cite{Bianchi23SofTossLearning}. Using a \ac{NN} to learn the direct model of the throwing task, accounting for non-linearities and delays typical of pneumatic soft robots, Bianchi et al. show how an \ac{RL} policy is able to effectively learn dynamic tasks even when the system is subject to large uncertainties.

\subsection{Automation of Heavy Machinery}
Many of the techniques developed in robotic manipulation have been transferred to large scale heavy machines. However, material handling automation has rarely been the subject of research, despite being a crucial component in any construction task since these machines are the most efficient ones to manage equipment and material transport.

Existing works primarily focus on construction, including trenching~\cite{Jud19AutonomousFreeForm},excavation~\cite{Egli22GeneralApproach,Terenzi23AutonomousExcavation,Jin23LearningExcavation}, or drywall stacking~\cite{Johns23FrameworkRobotic}.
Mining and earthwork, conceptually similar tasks, have also been studied~\cite{Zhang21AutonomousExcavator,Aoshima22ExaminingSimulationtorealitygap}.
Farming and forestry applications are presented in literature as well~\cite{Jelavic22RoboticPrecision, Andersson21ReinforcementLearning}.

Among the previously investigated approaches, learning-based methods~\cite{Egli22GeneralApproach, Lee22PrecisionMotion, Spinelli24ReinforcementLearning} have been proven suitable to learn and leverage complex machine dynamics while being robust to disturbances under different deployment scenarios. We decided to pursue the same direction to address the control of hydraulic machines with redundant and underactuated kinematics.

Recent advancements in material handling automation show preliminary success. \cite{Spinelli24ReinforcementLearning} developed an \ac{RL} controller able to achieve position control of large material handlers with speed and accuracy comparable to human drivers. 
Their methodology surpasses human capabilities in dampening gripper oscillations, thereby enhancing safety during repetitive working routines. 
However, this approach doesn't leverage the passive end-effector dynamics. While being necessary for accurate gripper positioning, this approach prevents the full utilization of the machinery's capabilities and is limiting operational efficiency.
Building upon similar \ac{RL} foundations, our work presents a learned controller that uses tool oscillation as an exploitable component to increase machine reach and cuts the time for deceleration. 

Similar ideas have been investigated in ~\cite{Andersson21ReinforcementLearning}, where \ac{RL} is used for training a log manipulation controller on an underactuated forestry crane. 
Their work, despite being tested only in simulation, shows how a learning-based approach can successfully exploit gripper oscillation to complete the log grasping task and even be tuned to reduce energy consumption.
\section{Methodology}
\label{sec:methodology}

\begin{figure}[t]
  \centering
  \includegraphics[width=\columnwidth]{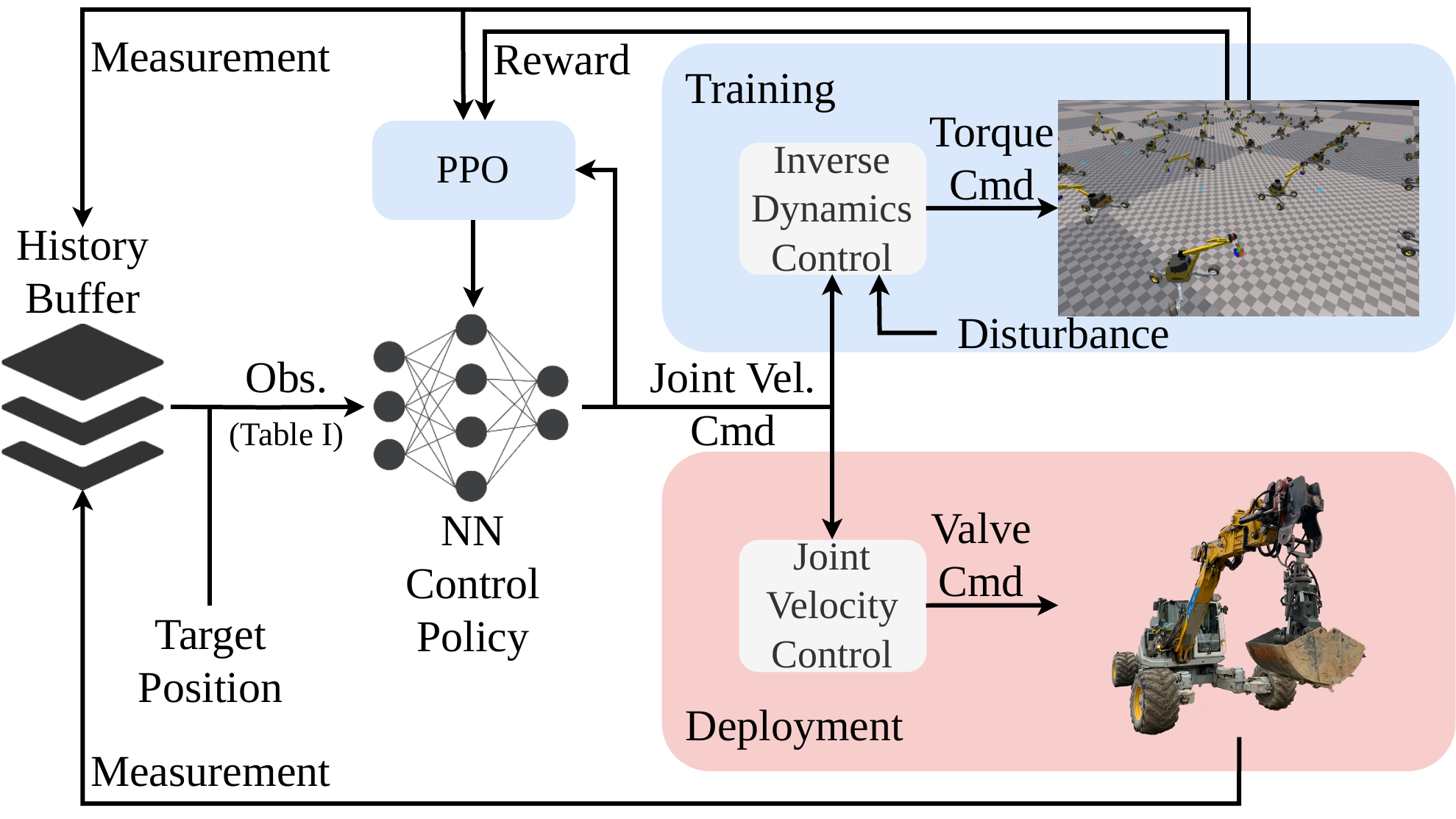}
  \caption{The pipeline for training and deployment of the controller. The learned throwing controller outputs joint velocity commands. During training, a torque-based simulator is used with a joint velocity controller. During deployment, a low-level controller replaces the joint velocity controller in simulation to command the hydraulic valves.}
  \label{fig:method_overview}
\end{figure}

We use \ac{RL} to train the throwing controllers purely in simulation and later deploy them on the real machine.
Due to the lack of fast parallelized simulation environments for hydraulic actuator dynamics, a torque-based simulator is used for training.
As shown in Fig. \ref{fig:method_overview}, a low-level joint velocity controller is introduced in the simulation environment, while in the real world a PID joint velocity controller with feed-forward compensation commands the proportional valves~\cite{Jud21HEAPAutonomous}.
As the rigid-body dynamics in the training environment can be obtained accurately, we first design the low-level controller in simulation to be accurate. For transfer robustness, we add artificial velocity command randomization during training.
This approach eliminates the need to simulate complex hydraulic system dynamics during training.
It significantly decreases training time and has been used in previous works that achieved successful sim-to-real transfers~\cite{Egli22SoilAdaptiveExcavation, Spinelli24ReinforcementLearning}.
This section describes the proposed training setup and necessary real-to-sim and sim-to-real steps for a successful real-world deployment.

We introduce two different control policies for the task of payload throwing.
The 3D policy utilizes all joints on the upper carriage of the machine.
Combining the motion of cabin turn, boom, dipper, and telescope as shown in \cref{fig:sys_overview}, can results in high end-effector velocity.
The usability of this controller is constrained by the availabe free space around the machine.
Therefore, we additionally introduce a 2D policy with the limitation of a fixed cabin turn joint.
In this case, only in-plane motions of boom, dipper, and telescope are performed for throwing.
Both controllers are trained to reach targets at different distances beyond the maximum conventional reach of the arm.

\subsection{System Description}
\label{sub:hardware}

\begin{figure}[h]
  \centering
  \includegraphics[trim={0 2cm 0 5cm},clip,width=\linewidth]{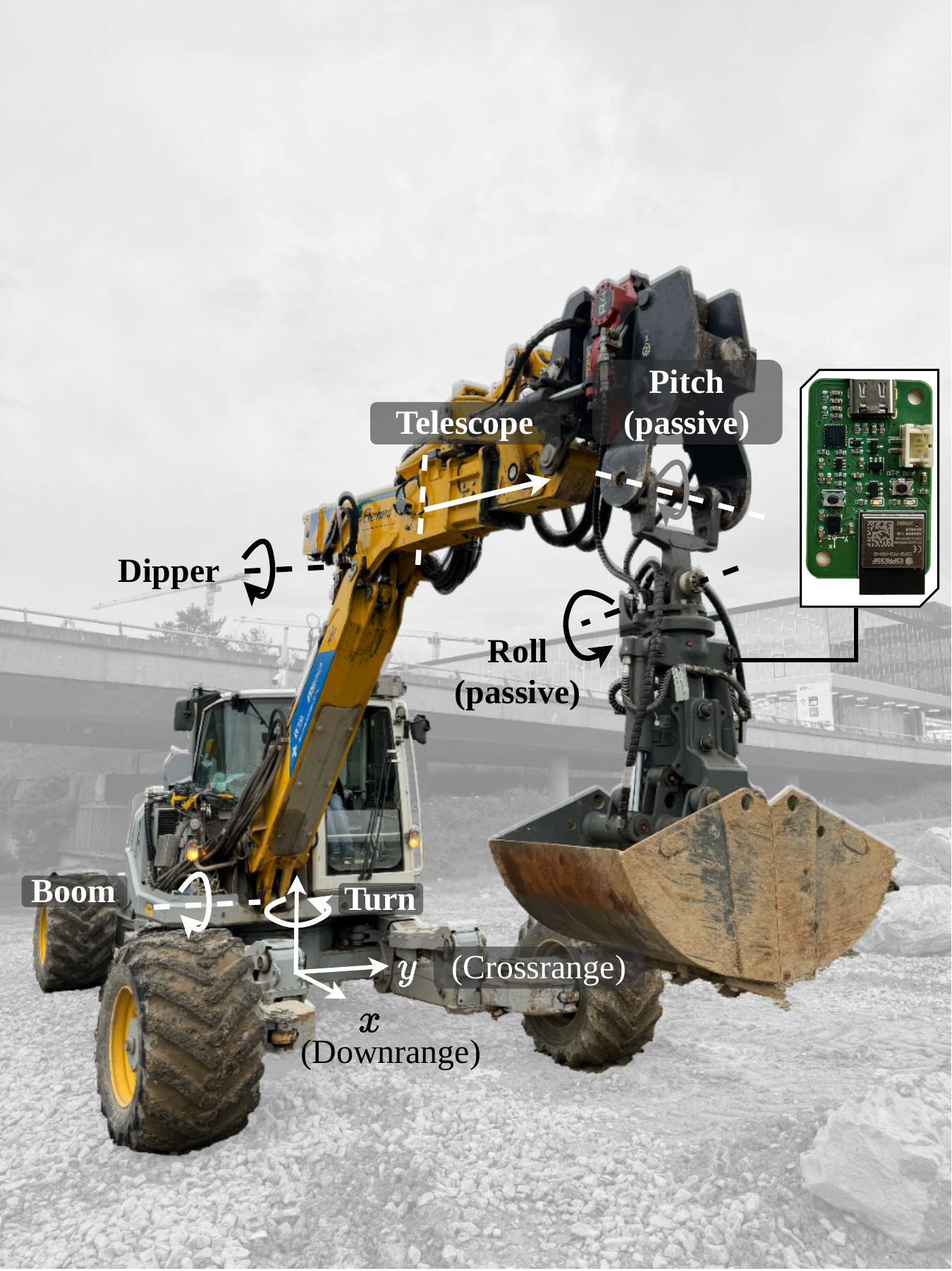}
  \caption{The material handling test platform is a 12-ton multipurpose excavator with a material handling gripper. A wireless \ac{IMU} module is attached to the gripper for state estimation of the passive joints. The \acp{DoF} considered in this work are marked in the picture.}
  \label{fig:sys_overview}
\end{figure}

The test platform used in this work is a modified version of the autonomous excavator HEAP~\cite{Jud21HEAPAutonomous}.
A material-handling gripper is installed, as shown in \cref{fig:sys_overview}.
The gripper has two passive \acp{DoF} around the pitch and roll axes.
As indicated in the figure, we only use four actuators on HEAP's upper body for the throwing task.

A small, battery-powered, wireless \ac{IMU} module was developed for the purpose of identifying gripper dynamics and passive joint state estimation.
The PCB carries an ST ASM330 automotive-grade \ac{IMU} and an ESP32 for signal filtering, pose estimation, and communication.
Rosserial tcp\footnote{\url{http://wiki.ros.org/rosserial}} is used to communicate the time-synchronized pose readings to the main computer on the excavator.

\subsection{Simulation for Training}
We built the simulation environment in Isaac Gym~\cite{Makoviychuk21IsaacGym}, a simulator that performs fast GPU-based physics simulation.
Assuming the throwing target is reachable without relocating the machine's chassis, we only simulate the six movable joints of HEAP.
In the training simulation environment, we implement the low-level joint velocity controller with an \ac{ID} approach.

During training, the excavator's active \acp{DoF} are initialized at random positions without initial velocity, while the tool is always aligned with gravity. 
The throwing target positions are also randomly sampled at ground level up to $45\%$ farther than the maximum reach of the arm. The horizontal distance from the target to the cabin turn center is constrained to have a minimum value so that targets leading to self-collisions are avoided.

During simulation, the controller provides joint velocity commands to the \ac{ID} controller at every step.
Payload release is handled by spawning a ball at the gripper's center with the same initial linear velocity.
Releasing the payload is delayed by a constant time that has been previously identified on the machine, as explained in \cref{sub:sim_to_real}. 
Subsequently, the ball follows a ballistic trajectory.
The release of the payload is only triggered once per rollout. 
Episodes terminate whenever a self- or ground collision is detected, or when the ball hits the ground.

\subsection{Observations and Actions}
The control policy is an \ac{MLP} network with hidden dimension size $\left[256, 128\right]$.
The input to the policy network mostly consists of the machine's proprioceptive information, including its joint positions, joint velocities, previous commands, and target position.
A full list of the 65-dimensional observation space for the 3D policy is given in \cref{tab:observation_action_space}.
Subscripts of symbols in the table indicate joint indices in a serial order of the kinematic chain.
Superscripts indicate the time step with $k$ being the latest step.
The joint states are fed to the policy with a history of three time steps, allowing the agent to infer the gripper dynamics and the behavior of the joint velocity controller.
We also provide the previous actions, encouraging the controller to learn consistent and smooth motions.
Similar observation vectors have been demonstrated helpful for successful deployment in the real world~\cite{Egli22GeneralApproach} in order to adapt to low bandwidths and large delays of hydraulic actuators.
Additionally, the measurements from a hydraulic machine are usually noisy and the history helps the policy to infer the real joint states.

To train the 2D control policy, we removed the cabin turn joint position, joint velocity, and previous command from the observation space, resulting in a 56-dimensional observation space. 
The same \ac{MLP} architecture is used for both policies. 

\renewcommand{\arraystretch}{1.2}
\begin{table}[h]
  \centering
  \caption{Observations of the Control Policy, 3D}
  \begin{tabular}{lcr}
    \toprule
    Observation & Notation & Dimension \\
    \midrule
    Joint positions & $\bm{q}_{1:6}^{k, k-1, k-2, k-3}$ & 24 \\
    Joint velocities & $\bm{\dot{q}}_{1:6}^{k, k-1, k-2, k-3}$ & 24 \\
    Previous commands & $\bm{u}_{1:5}^{k-1}$ & 5 \\
    Gripper center position & $\bm{r}_{EE}$ & 3 \\
    Gripper center velocity & $\bm{\dot{r}}_{EE}$ & 3 \\
    Gripper position error X-Y plane & $ err_{2D} $ & 1 \\
    Gripper position error 3D & $ err_{3D} $ & 1 \\
    Target position & $\bm{r}^{*}$ & 3 \\
    Gripper opened (binary) & $ i_{\text{opened}} $ & 1 \\
    \bottomrule
  \end{tabular}
  \label{tab:observation_action_space}
\end{table}

Our 3D throwing controller predicts a command vector $a$ in the following form:
\begin{equation}\label{eq:control_law}
    \bm{a} = \left[\hat{\dot{q}}_{1}, \hat{\dot{q}}_{2}, \hat{\dot{q}}_{3}, \hat{\dot{q}}_{4}, \hat{q}_{\text{release}}  \right]^T,
\end{equation}
while the 2D policy does not include the cabin turn command $\hat{\dot{q}}_{1}$.
In addition to the joint velocity references $\hat{\dot{q}}_{1:4}$, the control policy outputs a gripper opening command $\hat{q}_{release}$.
When the command value exceeds a predefined threshold $0.9$, it triggers the release of the payload.
Furthermore, this action is overwritten by a constant releasing command, with a $5 \%$ probability, when the gripper reaches a neighborhood of the target location with low speed.
This modification, despite violating the rigorous \ac{RL} update rule and could lead to slower policy updates, allows for easier and better exploration over the discrete action space.

\subsection{Reward and Terminations}
The per-step reward function is in the following form:
\begin{equation}
  r = c_1 r_{\Delta_{err}} + c_2 r_{err_{3D}} - c_3 r_{act_{diff}} - c_4 r_{act} \\
\end{equation}
with 
\begin{align*}
  r_{\Delta_{err}} &= \max{\left(0, err^*_{2D} - err_{2D}\right)}, \\
  &\quad + \max{\left(0, err^*_{3D} - err_{3D}\right)}, \\
  r_{err_{3D}} &= \exp{\left(-b_1 \left\lVert \bm{r}^* - \bm{r}_{ball} \right\rVert_2^2 \right)}, \\
  r_{act_{diff}} &= \left\lVert \bm{u}_{1:5} - \bm{u}_{1:5}^{k-1} \right\rVert_2^2, \\
  r_{act} &= \left\lVert \bm{u}_{1:5} \right\rVert_2^2 ,\ \text{if}\ i_{\text{opened}} = 1,
\end{align*}
where $err^*_{2D}$ and $err^*_{3D}$ are the minimum X-Y plane and 3D distances from the ball to the target position ever reached during each rollout. All weights $b_\cdot, c_\cdot$ are positive. The weights $c_1$ to $c_3$ were tuned independently for the two control policies due to the differences in action dimension and scale of position errors.
$r_{err_{3D}}$ and $r_{\Delta_{err}}$ are provided at each step to encourage transporting the ball towards the target position.
$r_{\Delta_{err}}$ is necessary because a high value of $r_{err_{3D}}$ can be attained by keeping the gripper next to the target position without releasing the ball.
With a higher weight on $r_{\Delta_{err}}$, such behavior can be avoided.
Over the entire process, the penalization term $r_{act_{diff}}$ incentivizes the policy to perform smooth motions.
A very small regularization term $r_{act}$ is used to prevent meaningless actions after the ball is released, preparing the machine for the next sequence.
A negative termination reward is given when the episode terminates due to self-collision, ground collision, or reaching joint limits. When the rollout terminates due to the ball dropping to the ground, a positive termination reward in the form of $\exp(b_2 err_{3D}^2)$ is given to the agent.

\subsection{Training}
The control policy is trained using the \ac{PPO} algorithm~\cite{Schulman17ProximalPolicy} with generalized advantage estimation\footnote[1]{The implementation can be found at \url{https://github.com/leggedrobotics/rsl_rl}}.
The simulator runs at \SI{100}{\hertz}, and the control frequency during training is \SI{12.5}{\hertz}.
We train with 8192 parallel environments and 4000 iterations with 32 control steps per iteration.
The policy training takes about 1.7 hours on a desktop computer with an Intel i9-13900K CPU and an NVIDIA RTX4090 GPU using 2.66 years of experience.

\subsection{Sim-to-Real Transfer}
\label{sub:sim_to_real}
Successful execution of the learned motions on hardware requires additional effort to bridge the sim-to-real gap.
Besides random noises added to the observation vector and disturbances in the low-level controller, we performed parameter identification on the hardware to remove the repeatable mismatches between the simulation and the real world.
Domain randomization was applied to the identified parameters to further robustify the policy against random noise and unmodeled dynamics.

\subsubsection{Actuation Delay}
On a hydraulic machine with proportional valves, the performance of actuator control is usually limited by large delays and low bandwidth.
The opening delay of the gripper has been experimentally identified by dropping the wireless IMU from the gripper and recording the time from command to free-fall. 
The identified actuation delay is \SI{258}{\ms} with a standard deviation of \SI{15}{\ms} after 39 releases.

\subsubsection{Passive Joint Friction}
\begin{figure}[t]
    \centering
    \includegraphics[width = 0.8\linewidth]{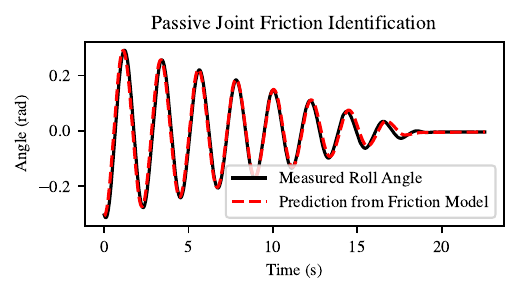}
    \caption{Identification of the passive joint friction in pitch / Y direction. Comparison of simulated oscillation and measurements from the \ac{IMU}.}
    \label{fig:plot_fric_model}
\end{figure}

Having a good model of friction and damping for the passive joints is critical, as the controller has to move the arm accordingly to excite these joints.
The free motion of the gripper is recorded with the wireless IMU, and the data is fitted to the friction model outlined in \eqref{eq:friction_model}. $\Ddot{\Theta}_{\text{fric}}$ represents the joint acceleration caused by friction and damping effects, $\upsilon$ the damping coefficient, $\dot{\Theta}$ the joint velocity and $\eta$ the velocity independent friction component.
A damping coefficient $\upsilon$ of 0.03 and a friction coefficient $\eta$ of 0.1 are identified for our gripper, indicating a dominant velocity-independent friction part.
Both passive axes behave similarly.
\Cref{fig:plot_fric_model} compares the measured motion of a passive joint when releasing from an initial angle and the predicted motion from the identified parameters.
The coinciding curves indicate a small sim-to-real gap with the identified model.

\begin{align}
    \Ddot{\Theta}_{\text{fric}} &= \upsilon \cdot \dot{\Theta} + \eta \cdot \text{sign}(\dot{\Theta})
    \label{eq:friction_model}
\end{align}

\section{Experiments}
\subsection{Experiments in Simulation}
\label{sec:simulation}
The quantitative throwing performance was primarily evaluated in simulation to gather relevant statistics.
Later, a smaller number of real-world experiments was carried out to validate the result.
To avoid benchmarking the trained policies on the seen environment and dynamics, the evaluation was performed in a different simulator, Gazebo. 

The Gazebo simulator uses a different, PID-based joint controller which represents the real world joint dynamics with higher fidelity compared to Isaac Gym.
Even though this sim-to-sim transfer can only partially show the parameter change experienced during real-world deployment, it still challenges the robustness of the controller and allows for a training-independent performance evaluation.

The two presented controllers are independently evaluated for precision and later compared for practicality.

\subsubsection{Gazebo Simulation Setup}
To benchmark the controllers, the Gazebo simulator setup from \cite{Jud21HEAPAutonomous} is used.
The passive gripper was added to the simulation environment using the same identified friction and damping as described in \ref{sub:sim_to_real}.

For each evaluation run, the machine's arm joints get randomly re-initialized at collision-free positions.
This ensures, that the measured performance is independent of the initial arm configuration. 
Target points get sampled along the downrange axis of the machine at a distance from the cabin.
We focus on distances at the border and outside of the arm's static reach as the throwing motion is particularly useful for these targets.
Due to the symmetry of the task and training of arbitrary targets, sampling only along the downrange axis is sufficient even for the 3D policy evaluation.

\begin{figure}
    \centering
    \includegraphics[width = \linewidth]{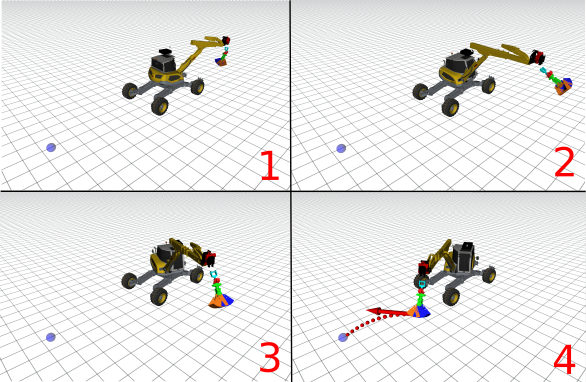}
    \caption{Visualization of the evaluation in sim. 3D policy throwing at a target at \SI{9.5}{\m} distance with a static reach of \SI{7.5}{\m}. Target points are indicated in blue, payload trajectory in red with the initial velocity vector on release.}
    \label{fig:rviz_throwing_trajectory}
\end{figure}

\Cref{fig:rviz_throwing_trajectory} shows the evaluation environment with a randomly selected target point and throwing trajectory from the 3D policy.
Here, the cabin turn is initialized nearly \SI{180}{\degree} away from the target.
The 3D controller turns the cabin joint rapidly to swing up the passive joints to generate a high radial velocity for throwing at the target.

\subsubsection{2D Policy}
\begin{figure*}[!h]
    \centering
    \includegraphics[width = 0.8\linewidth]{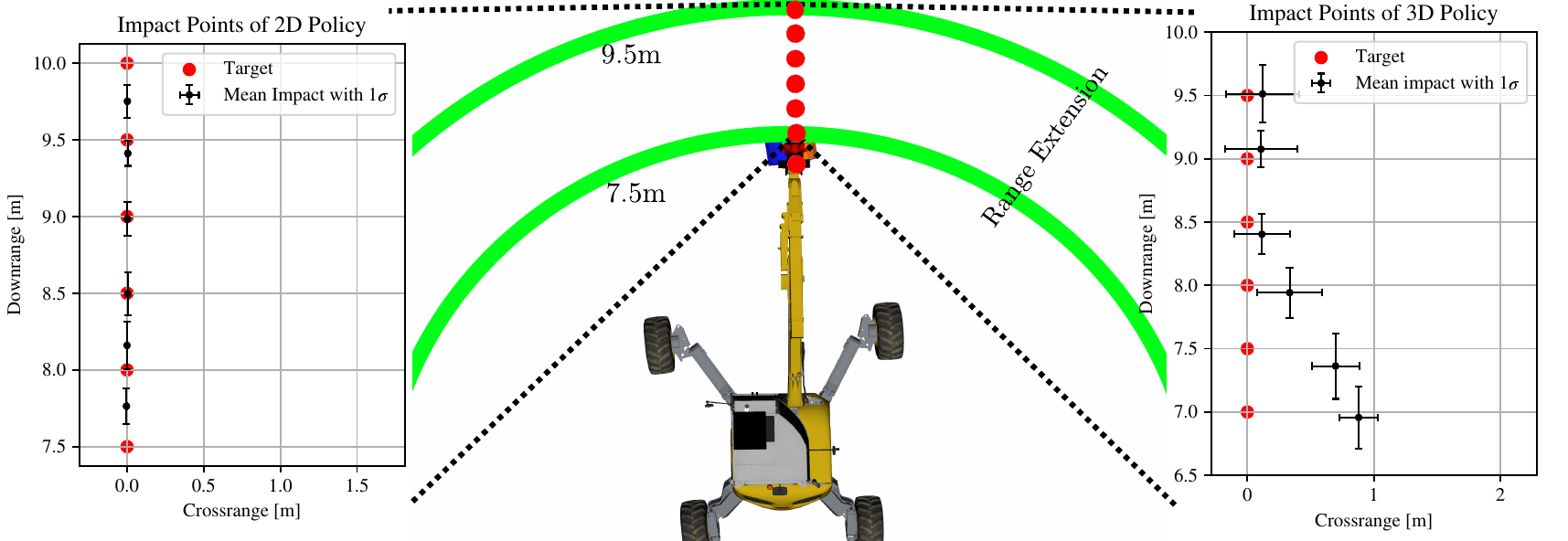}
    \caption{Visualization of the achieved extended range through agile material handling. The plot on the left side shows the evaluation target accuracy results from simulation for the 2D policy. On the right, the equivalent plot for the 3D policy is shown. }
    \label{fig:rangeExtension}
\end{figure*}

To evaluate the 2D Policy, the arm of the excavator is aligned with the downrange axis.
Target distances from \SI{7.5}{\m} to \SI{10}{\m} in steps of \SI{0.5}{\m} are evaluated.
Each target distance is repeated 200 times from different starting configurations.
The left plot in \cref{fig:rangeExtension} shows the achieved range accuracy with mean impact point and standard deviation.

It is visible, that the controller achieves good accuracy and repeatability up to \SI{9.5}{\m}.
Targets close to the conventional reach of the arm are overshot but the spread of impact points remains low with a $1\sigma$ spread of \SI{11}{\cm} and \SI{15}{\cm} for the three close targets.

The learned motion raises the arm to a certain height and uses the stored potential energy to swing the passive pitch joint of the gripper. A sudden stop of the boom joint accelerates the passive joint and the gripper is opened at the correct time. \cref{fig:throwing_trajectory} depicts the motion of the 2D policy throwing a ball into a bucket.

\subsubsection{3D Policy}
Adding cabin turn motion enables the controller to build up kinetic energy without the need to lift the gripper.
As shown in \cref{fig:rviz_throwing_trajectory}, the learned motion accelerates the cabin joint and releases the payload with significant tangential velocity.
This way, targets not limited to the downrange axis can be reached.

Adding cabin motion simultaneously spreads the impacts along the crossrange axis.
\cref{fig:rangeExtension} on the right plot shows the evaluation from Gazebo for targets between \SI{7.0}{\m} and \SI{9.5}{\m}.
In this evaluation, the cabin motion is clockwise for all samples.
Impacts for targets up to \SI{8.0}{\m} show a distinct offset in the direction of motion. 
The release point for these throws is close to the point itself which increases the sensitivity to inaccurate release timing.
Further targets show better precision with an overall similar accuracy of approximately \SI{30}{\cm} standard deviation in both directions for all target distances.

\subsubsection{Evaluation Conclusion}
The two learned motions target different initial configurations.
While the 2D policy is more precise in all reached distances, it requires previous alignment with the target.
Arbitrary starting and target positions can directly be handled by the learned 3D policy.
The 3D motion is much more dynamic but might pose a significant risk for everything close to the machine.
By using either presented controller, the effective dumping range of the machine can be significantly increased from \SI{7.5}{m} to \SI{9.5}{\m} as visualized in \cref{fig:rangeExtension}

\subsection{Real World Experiments}
\label{sec:experiment}
Real-world experiments are conducted to verify the accuracy of the simulation results.
Both policies are evaluated with a target at \SI{9}{m} downrange and \SI{0}{m} crossrange.
Each policy is tested 10 times with randomized initial positions.
As a payload, a standard size 7 basketball is used for good visibility.
The 2D policy validation setup can be seen in \cref{fig:throwing_trajectory}.
\begin{figure}
    \centering
    \includegraphics[width = \linewidth]{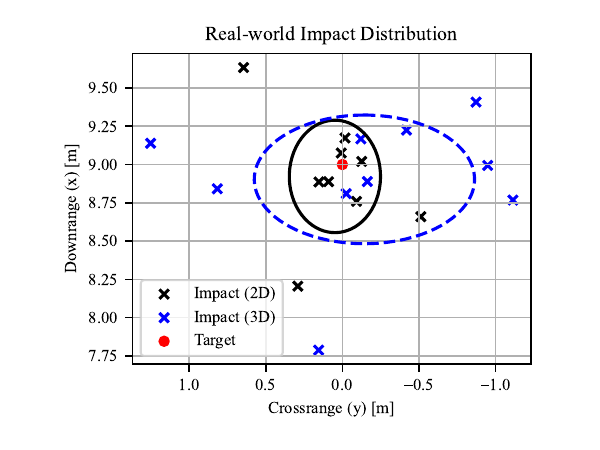}
    \caption{Results of the real-world evaluation.
    The crosses represent the impact locations for the 2D and 3D policies. The ellipsoids mark one standard deviation. The target for both policies is denoted by the red dot.}
    \label{fig:results_realword}
\end{figure}
\Cref{fig:results_realword} shows the results of the experiments.
Both policies can hit a downrange distance of \SI{9}{m} with a mean downrange impact location of \SI{8.92}{m} for the 2D policy, and \SI{8.9}{m} for the 3D policy.
As expected, the 2D policy shows superior crossrange accuracy, showing a mean crossrange error of \SI{0.05}{m} compared to the \SI{-0.14}{m} achieved by the 3D policy.

The 2D policy shows standard deviations of \SI{0.37}{m} and \SI{0.3}{m} in the downrange and crossrange directions respectively.
The observed crossrange error is caused by the cabin slowly drifting during the evaluation run.
Since the policy does not command the cabin turn joint, vibrations during the individual throws slowly change the cabin orientation.
Furthermore, small differences in the payload grasp, such as a slight off-center grasp of the basketball, also contributes to this error.

While the 3D policy achieves a similar downrange precision with a standard deviation of \SI{0.42}{m}, the crossrange precision suffers, showing a standard deviation of \SI{0.72}{m}.
As the 3D policy is making full use of the cabin turn joint, this error is expected. Even small differences in the release timing will have a significant effect on the crossrange impact location due to the additional velocity gained by the rotation of the cabin.
Differences in the grasp strength of the payload may also introduce an additional release delay, which will have a larger impact on the crossrange error using the 3D policy.
Since the achieved standard deviation is of similar size to the bucket itself, the expected accuracy when handling bulk material will not be degraded by the performance of our controller.
\addtolength{\textheight}{-1cm}

\section{Conclusion and future work}
\label{sec:conclusion}
In this work we show a controller trained with \ac{RL} that performs dynamic throwing with automated heavy machinery.
Through training in simulation, the obtained control policies learned to exploit the underactuated kinematics of the machine to perform dynamic throws.
The controllers were successfully deployed in a different simulation environment and in the real world.
The learned controllers perform throwing motions in unconstrained environments which extend the reach of the machine at a precision similar to the size of the gripper itself. This accuracy renders our controller not only suitable for dumping material in free space, but also to accomplish common logistic tasks on a construction site. 
Such throwing controllers could remove the need to dampen the passive joints before material release, leading to a potential boost in efficiency by reducing cycle times.
We believe these benefits will significantly improve the productivity of relevant industries and lay the path to fully autonomous machines.

In upcoming research we plan to extend the presented method in order to tackle more realistic scenarios.
We focus on showing the capability to throw with the underactuated cranes, but for practical applications, the controller should allow the user to trade-off between more dynamic throws and more accurate ones.
Another interesting challenge will be the addition of throwing to 3D targets, e.g., throwing bulk material onto the top of a pile.
We aim to develop a simulation that mimics real-world handling of bulk materials, focusing on the efficient deposition of granular material. To enhance this process, we will integrate the presented method with a sophisticated planning algorithm that selects the most effective points for material placement, optimizing the dumping strategy.







\section*{ACKNOWLEDGMENT}
The authors would like to thank Edo Jelavic and Andreas Dietsche for their valuable support in setting up the simulation and hardware.



\bibliographystyle{bibliography/IEEEtranN}
\bibliography{bibliography/references}

\end{document}